# A Binary PSO Based Ensemble Under-Sampling Model for Rebalancing Imbalanced Training Data

Jinyan Li[1], Yaoyang Wu[1, *], Simon Fong[1], Antonio J. Tallón-Ballesteros[2], Xin-she Yang[3], Sabah Mohammed[4]

* Corresponding author: mb65519@connect.um.edu.mo

[1]Department of Computer and Information Science, University of Macau, Taipa, Macau SAR, China

Email: yb47432@connect.um.edu.mo, mb65519@connect.um.edu.mo, ccfong@umac.mo,

[2]Department of Languages and Computer Systems,University of Seville, Seville, Spain

Email: atallon@us.es

[3]Design Engineering and Math, School of Science & Technology, Middlesex University, London, UK

Email: X.Yang@mdx.ac.uk

[4]Department of Computer Science, Lakehead University, Thunder Bay, Canada

Email: sabah.mohammed@lakeheadu.ca

## Abstract

Ensemble technique and Under-sampling technique are both effective tools for resolvig imbalanced dataset classification problems, which commonly denotes the quantitative imbalance of a binary class dataset where minority class is the target class. In this paper, a novel ensemble method absorbing the advantages of both ensemble learning for biasing classifiers and a new evolutional under-sampling method is proposed. The under-sampling method is named Binary PSO instance selection, it gathers with ensemble classifiers to find the most suitable length and combination of majority class samples to build a new dataset with minority class samples. The proposed method adopts multi-objective strategy, objectives of this method is to improve the performances of imbalanced classification and to guarantee the maximum integrity of the original dataset. We examined our proposed method, the Binary PSO instance selection by comparing its performance of processing imbalanced dataset with several other conventional basic ensemble methods. Experiments are also conducted with Binary PSO instance selection wrapping with ensemble classifiers for further improvement of imbalanced classification. Based on comaprison experiments, our proposed methods outperform single ensemble methods , state-of-the-art under-sampling methods as well as their combinations with traditional PSO instance selection algorithm.

**Keywords**: Imbalanced classification, Ensemble, Under-sampling, Binary PSO, multi-objective,Intergrity.

## 1. Introduction

Classification is a main task in data mining and machine learning. Classification algorithms build classification models through training datasets, where models are used for prediction of unknown class samples. Nowadays, many classifiers could obtain significant results of classifying balanced distributed datasets. However, there are many datasets in real life that are imbalanced, which conventional classifiers might not be able to provide satisfactory performance on imbalanced classifications. Essentially, in imbalanced datasets, the quantity of samples from some classes is far more than others, these type of classes are usually called majority class, and the alternative is minority class. For research purposes, the minority classes are usually the ones that we are interested in, and binary class imbalanced datasets are the most commonly processed type of imbalanced classification. For example, satellite radar position [22], telecom customer problem [13], fraud cases [16], network intrusion [31] and detection of biological datasets [38], the interesting class has only very few samples.

The reason why conventional classifiers might not be suitable for imbalanced classifications is that most classifiers assume the classified dataset is balanced to seek the maximum accuracy of classification model. However, the prediction effect of minority class is poor since minority class is scarce. For instance, in a binary class dataset, the minority class samples accounts for 1% of the total and the rest samples belong to majority class, in the process of classification, classifiers are commonly biased towards majority class and they neglect minority class, the accuracy of this training classification model can be as high as 99%. Yet this classification model is useless for identifying and predicting our interested minority class samples. This phenomenon presents a based problem of imbalanced classification, and this high accuracy is called high pseudo-accuracy [34]. Therefore, the robustness of imbalanced classification model for the meaningful minority class samples is very low, which can be reflected by some metrics, like Kappa statistics, G-mean, BER, etc.

In this paper, we propose a new evolutional under-sampling method called Binary PSO Instance selection, and combine it with ensemble methods to solve imbalanced classification problems. Under-sampling technique reduces the number of majority class samples to diminish the imbalanced ratio of the original dataset and improve performance of imbalanced classification. If there are $N$ samples of majority class, then there are $2^N$ combinations to structure the candidate solutions. Moreover, under-sampling method needs to consider both the length and elements of these candidate solutions. That means the computational cost is big. In addition, ensemble methods could slightly change the tendency of conventional classifiers therefore promoting the performance of the classification model. That depends on the effect of ensemble methods influenced by the parameter setting and the highly imbalanced distribution of the original dataset. Instance selection is necessary for removing some possible gibberish in original dataset in order to improve the performance of the classification model and diminish the imbalanced ratio. However, we also need to respect the original data in data science, in order to reflect the objective results. Thereby, Binary PSO Instance selection is designed to increase the performance of imbalanced classification and ensure the maximum integrity of the original dataset, simultaneously. It was implemented through controlling multi-objective. Furthermore, ensemble methods could improve the imbalanced classification without changing the original dataset. Hence, finally, wrapping Binary PSO Instance selection with ensemble classifiers is a useful method capable of forming a higher performance of classification model while obtaining a most integral dataset possible along the process.

The paper is organized as follows. Section 2 reviews the previous methods that are used to solve imbalanced classification problems. In Section 3, details and process of the proposed method solving imbalanced dataset is described. Section 4 contains the benchmark dataset description, experimental procedure and result analysis. Section 5 summarizes this paper.

## 2. Related works

Imbalanced classification problem is a popular topic in the in data mining, machine learning and pattern recognition fields. There are many leading conferences held special workshop for discussion and studying for this problem, like in ACM SIGKDD 2004[6], AAAI 2000 [21], ICML 2003 [44][10], etc. Present days, the researches for solving imbalanced classification can be roughly recognized as two categories: data level and algorithm level. Previous researcher proposed that there are four main factors for tackling imbalanced classification problems, they are: training set size, class priors, cost of errors in different classes and placement of decision boundaries [5]. The data level aims to reduce the imbalanced ratio of imbalanced classification model by adjusting the distribution of samples in dataset. Another level of the algorithm makes the classifier more inclined towards the minority class through modifying conventional classifier.

From the design of most conventional classifiers, previous researchers found that the performance of balanced dataset is better than that of imbalanced classification [12]. Therefore, people proposed many methods for rebalancing the imbalanced dataset, in order to change the distribution of samples and rebalance the imbalanced dataset. Over-sampling and down-sampling respectively increase the number of minority class samples and decrease the number of majority class samples. Random over-sampling means randomly repeat minority class samples to increase the number of minority class samples, but this method will easily cause over-fitting [7]. Chawla proposed synthetic minority over-sampling technique (SMOTE), which is the most widely and effectively used over-sampling method, it synthetics new minority samples through learning from several neighbors in the same class of each minority class sample, in order to generate minority samples and rebalance the imbalanced dataset. Although over-sampling technique is able to reduce the imbalanced ratio, the original minority class samples may be diluted by a large amount of synthetic samples. Down-sampling discards a part of majority class samples to rebalance the dataset

[40]. Random down-sampling could cause the loss of some valuable and characteristic samples. Balance Cascade [29] is a classical under-sampling method. Through iteration strategy, it removes the useless majority class samples step by step.

The algorithm level contains two main approaches to improve imbalanced classification, cost-sensitive learning and ensemble learning. In the classification process, they make the base classifiers favor more the minority class samples than majority class ones through assigning different weights or voting or iteration.

As we know, in most cases, minority class samples are our targets to explore and study. Therefore, it is more valuable to correctly identify the minority class samples than the majority class samples. In other words, it would cost more for misclassifying minority class samples. Hence, it is the basic idea of cost-sensitive learning [11], which assigns different costs of misclassified classes. For example, in binary class imbalanced dataset, assuming negative is the minority class and the cost of misclassified minority class samples is higher. Therefore, in the training of classifier, the classifier will be forced to have a higher recognition rate for negative class samples since there will be greater punishment for misclassified negative class. The paper will mention confusion matrix and the cost matrix in the following section and give an example to introduce how the cost-sensitive learning is achieved to change the tendency of classifiers.

The basic idea behind the ensemble learning is that the algorithm will get a number of base classifiers from the training set, and then it uses some ensemble techniques to integrate them to improve the performance of classification. Bagging [39], boosting [18], random forest [8] are the most commonly used methods. Bagging improves the performance of classification through the vote of several single classifiers, which classifies the re-sampled (with replay) datasets from the original dataset. Its final results are combined by each sample which gets the most votes. Boosting methods are the most popular ensemble methods, its implementation is a process of iteration. Adaptive Boosting (AdaBoosting) is the representative in the family of boosting methods [41]. It adaptively changes the distribution of the training sample by assigning different and vibrational weights to each sample in iteration. We will introduce this algorithms in the next section in detail.

AdaBoosting and cost-sensitive learning were combined by some researchers to build AdaCost [15] series algorithms, AdaC1, AdaC2, and AdaC3 [42]. This kind of algorithm absorbs the benefits of Adaboosting and cost-sensitive learning. Moreover, SMOTEBoost algorithm [9] combines SMOTE method with boost method further improves the performance of imbalanced classification. It uses SMOTE to synthetic minority class samples in the iteration of AdaBoosting, in order to make the sub-classifiers pay more attention to minority class samples. The Support Vector Machines (SVM) [3] and feature selection [4] are also helpful for tackling class imbalance problem. Moreover, people also adopted some evolutionary algorithms to tackle imbalanced problem previously [20].

## 3. Methodology

In this section, we will describe the proposed new ensemble under-sampling method. Ensemble and Under-sampling are effective techniques for tackling imbalanced dataset classification problem, which commonly denotes the quantitative imbalance of a binary class dataset where minority class is the target class. Here, we propose a new ensemble method which combines the benefits of both ensemble methods for biasing classifiers and a new evolutional under-sampling method. Therefore, the "ensemble" has two meanings in our paper, the first meaning is the ensemble techniques, like bagging, boosting and stacking; the second implication is the proposed method binds previous ensemble techniques and undersampling techniques. The under-sampling method is named Binary PSO instance selection, it gathers with ensemble classifiers to find the most suitable length and combination of majority class samples to build a new dataset with minority class samples. The proposed method adopts multi-objective strategy, which simultaneously improves the performances of imbalanced classification and guarantees the maximum integrity of the original dataset. We examine the effect of Binary PSO instance selection by comparing the performance of processing imbalanced dataset with several other conventional basic ensemble methods. In the next step, Binary PSO instance selection is wrapped with ensemble classifiers for further tackling imbalanced classification.

### 3.1 Particle Swarm Optimization

Particle Swarm Optimization (PSO) [24] [45] is a widely used meta-heuristic algorithm which imitates the feeding process of birds. It has the advantages of easy implementation, faster convergence and fewer parameters. Moreover, since the simple requirement of the objective function and constraint conditions, it offers new solution and approach to solve non-linear and NP hard problems in different fields [35] [1].

Pseudo code of PSO:
1. **For** each particle
2.    Initialize particle and parameters
3. **End**
4. **While** maximum iterations or the termination mechanism is not satisfied.
5.    **For** each particle
6.      Calculate and update particle velocity and position as equation (2) and (3)
7.    **End**
8.    **For** each particle
9.      Calculation of fitness function
10.      **If** the fitness value is better than the best fitness value (pBest) in history
11.        **Do** current fitness value represent the older pBest to be the new pBest
12.      **End**
13.    **End**
14.    Selected the gBest whose fitness value is the best in the population.
15. **End**

Above-mentioned pseudo code describe the process of PSO. Assuming there is a population $X = (X_1, X_2, \ldots, X_n)$ which is grouped by $n$ particles in $D$ dimension search space, the $i^{th}$ particle in this space is expressed as a vector $X_i$ with $d$ dimension, $X_i = (x_{i1}, x_{i2}, \ldots, x_{id})^T$, and the position of the $i^{th}$ particle in the search space represents a potential solution. As the objective function, the program can calculate the corresponding fitness of position $X_i$ of each particle, where the speed of the $i^{th}$ particle is $V_i = (V_{i1}, V_{i2}, \ldots, V_{id})^T$, the extremum value of each agent is $P_i = (P_{i1}, P_{i2}, \ldots, P_{id})^T$ and the extremum of the population is $P_g = (P_{g1}, P_{g2}, \ldots, P_{gd})^T$. In the process of iteration, the extremum value of each agent and the population will update their position and speed. Equations (2) and (3) show the mathematical process as follows:

$$V_{id}^{k+1} = \omega V_{id}^k + c_1 r_1 (P_{id}^k - X_{id}^k) + c_2 r_2 (P_{gd}^k - X_{id}^k), \quad (2)$$

$$X_{id}^{k+1} = X_{id}^k + V_{id}^{k+1}. \quad (3)$$

In the Equation (2), $\omega$ is inertia weight and a nonnegative number; $d = 1, 2, \ldots, D$; $i = 1, 2, \ldots, n$; $k$ is the current iteration time; $c_1$ and $c_2$ are non-negative constants as the velocity factor, $r_1$ and $r_2$ are random values between 0 to 1 and $V_{id}$ is the particle speed.

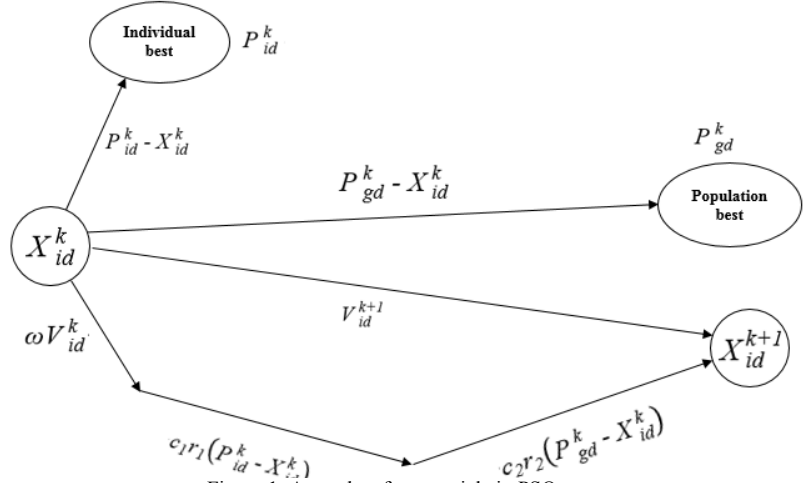
Figure 1. An update for a particle in PSO

$\omega$ controls the size of search space, and they are proportional relationship. Figure 1 shows the update of each particle. In the early stage of evolution, the position and velocity of each particle are randomly initialized, the population best $P_g$ attracts other individuals in the population, enabling them the other particles are rapidly converges to the global optimal region. Hence, PSO has strong global search ability, and the convergence speed is faster in the early evolution.

### 3.2. Binary Particle Swarm Optimization Instance selection for Multi-objective problem.

### 3.2.1. Swarm Instance Selection.

As we know, under-sampling for majority class is a useful method to solve imbalanced data classification problem. Furthermore, it is generally known that in the process of data collection, bad or error samples in a dataset are inevitable. Consequently, data cleaning of instance selection is essential. Swarm Instance selection for majority class samples is a kind of under-sampling method. There are non-linear relationships between different groupings of majority class and minority class samples, and Swarm instance selection adopts wrapper strategy to find the best combinations of majority class and minority class with the best classification results. Wrapper approach [32] is a commonly used method in evolutionary computation [25] [14]. It uses the selected solution to directly train the machine learning algorithm and evaluate the performance of the selected solution through testing the corresponding machine learning algorithm. Therefore, the effect of the wrapper approach is affected by the chosen machine learning algorithm. Wrapping swarm intelligence algorithms and machine learning algorithms (classifier) are able to obtain a significant solution.

Figure 2 presents the concept of swarm instance selection for majority class to tackle the imbalanced classification problem. As mentioned above, if the number of majority class is N, there are 2N candidate solutions. That means it is an N-P problem and brute-force is not achievable. Therefore, we choose swarm intelligence algorithms to search the optimal solution. The original imbalanced dataset is divided into two parts, one is minority class samples and the other is majority class samples for selection. For each particle we get a sub-majority class set, which will be gathered with original minority class samples to build a new dataset, then the classification performances of these new datasets will be tested. The population will move towards the global optimal, which is the combination of selected majority class samples and original minority class samples with maximum performance from the wrapped machine learning algorithm.

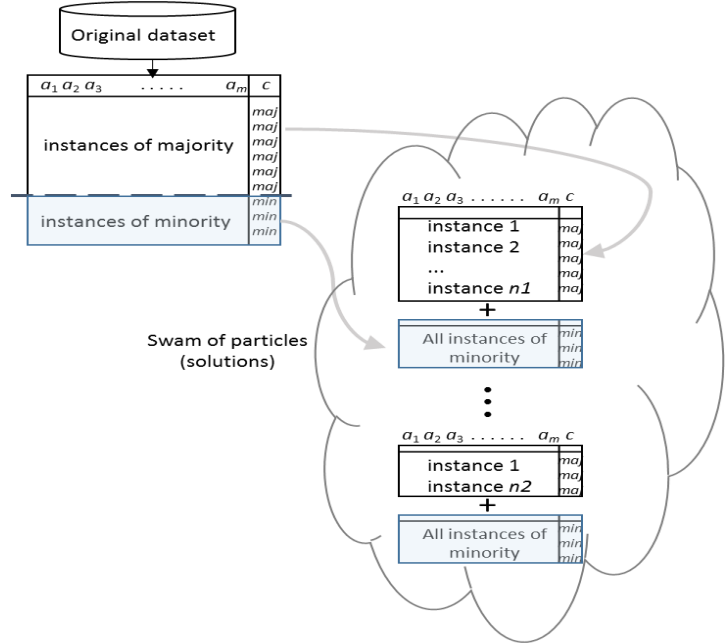

Figure 2. Principle of Swarm instance selection for majority class samples in imbalanced classification

### 3.2.2. Binary Particle Swarm Optimization Instance Selection

The concept of Binary Particle Swarm Optimization (BPSO) was proposed by Kenny and Eberhart [23], it is used to discretely handle discrete binary optimization problems. Based on PSO, the movement track and velocity of each particle are defined by probability, which is the probability of 0 or 1 in each particle's position and velocity in the process of iteration. They used new Equation (3) to replace Equation (2):

$$X_{id}^{k+1} = \begin{cases} 1, & rand() \leq S(V_{id}^{k+1}) \\ 0, & \text{otherwise} \end{cases}, \tag{4}$$

rand() is a random number in the open interval of 0 and 1. $S(V_{id}^{k+1})$ is a sigmoid function:

$$S(V_{id}^{k+1}) = \frac{1}{1+\exp(V_{id}^{k+1})}. \tag{5}$$

There is no doubt that swarm instance selection is a typical discrete binary optimization method. With traditional swarm instance selection approaches, one needs to consider both the length and elements of a selected majority class set simultaneously. While our proposed BPSO integrates these two parts through coding for particles. In this paper, the proposed BPSO inherited the idea of above method from Kenny and Eberhart [23] that the position of each particle can be given in binary form (0 or 1). , However, the new proposed BPSO in this paper introduces the instance selection process into a binary optimization problem. The numbers of majority class instances stand for the dimensions. This means that in our BPSO instance selection, the position of each individual particle can be given in binary form (0 or 1), which adequately reflects the straightforward 'yes/no' choice of whether a majority class sample should be selected. The scope of a position is from -0.5 to 1.5. Then, Equation (6) is used to calculate the binary value of the position [27]. It uses round function to simplify original BPSO.

$$X_{id}^{k+1} = \begin{cases} round(X_{id}^{k+1})=1 & 0.5 \leq X_{id}^{k+1} < 1.5 \\ round(X_{id}^{k+1})=0 & -0.5 \leq X_{id}^{k+1} < 0.5 \end{cases}, \tag{6}$$

Where the *round()* function calculates the binary value $X_{id}^{k+1}$ of the corresponding position to achieve the binary optimisation operation. Based on original PSO, Equation (6) is the follow set of Equation (3), however, both the position and velocity are a $1 \times N$ matrix, N is the number of majority class.

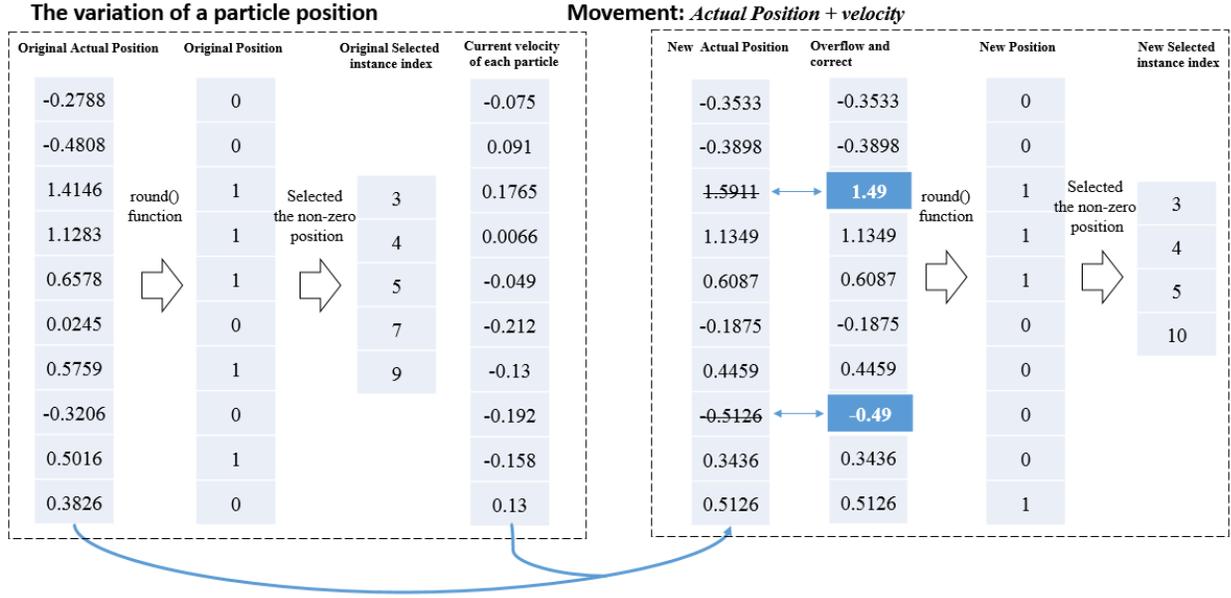

Figure 3. The variation of a particle of BPSO in iteration

Figure 3 presents a particle's movement in an iteration of a BPSO instance selection. The BPSO instance selection can be regarded as a high-dimensional function optimisation problem, where in the values of the independent variables are 0 or 1. In addition, values of 0 and 1 can also be given to dependent variables calculated by the rounding function, where independent variables can be assigned from -0.49 to 1.49. The step size of each position is a very small value in a fixed range. The motivation of the rounding function and the activity interval is aiming at expanding the searching space between 0 and 1. So that the position of particles can be more meticulous and precise. The classical definition of instance selection is selecting a sub-dataset *d* with *f* instance from the primary dataset *D* with *F* instance $f \leq F$, where d has the optimal performance in all of the sub-datasets with f instances from the primary dataset. Thus, we know that the value of *f* is a defined value in this definition, and while it should be a variable, it means that algorithms should find the optimal length with an optimal combination. The BPSO instance selection resolve this problem to obtain the optimal majority instance set using a similar method of function optimisation.

### 3.2.3. Multi-objective problem.

As mentioned, because of the imbalanced distribution of classes in imbalanced dataset, accuracy lost its effectiveness for evaluating imbalanced classification model. If the wrapper approach adopts accuracy as the fitness function, the high pseudo-accuracy from base learner will not be able to truly reflect the result of classification model to swarm. The binary confusion matrix [34] offers basic elements for calculating all metrics of classification model. Assuming negative class (N) is minority class, since the quantity of negative class samples is of low proportion in the dataset, the classifier would highly likely misclassify most, if not all of them into the wrong classes. That means if we use an all-negative class dataset as a testing dataset, the credibility of the trained classification model will be extremely low, because the classifier is under-trained with the minority class data. Therefore even the classification result presents a high accuracy of the model, it will all be meaningless when it comes to classifying imbalanced datasets. Equation (7) presents the mathematical formula to calculate a newly defined accuracy of classification model.

$$Accuracy = \frac{True\ Positive + Ture\ Negative}{P+N} \tag{7}$$

$$Sensitivity = TPR = \frac{True\ Positive}{P} \tag{8}$$

$$Specificity = TNR = \frac{Ture\ Negative}{N} \tag{9}$$

Sensitivity and Specificity respectively corresponding to the true positive rate and true negative rate. Assuming negative class (N) is minority class. True positive means majority class samples are correctly identified as majority class and true negative means minority class samples are correctly identified as minority class. Sensitivity and Specificity respectively refers to the test's ability to correctly detect the samples in majority class and minority class, they can be expressed as Equation (8) and (9). In other words, Sensitivity is high and specificity is very low in conventional imbalanced classification. However, Sensitivity and Specificity justly reflect the correctness of the classifications for both class samples. So we use the product of Sensitivity and Specificity as our first objective and fitness function in the search process of particles, in order to pursue the high results of Sensitivity and Specificity, synchronously. Figure 4 , an imbalanced dataset is used as an example and it used random under-sampling methods with different under-sampling rate of majority class samples to demonstrate the Snapshot of fluctuating values of TPR, TNP, TPR*TNR, Integrity of Majority class and Imbalanced ratio(min/maj) during random under-sampling with 20 different under-sampling rates. The performance is evaluated based on the mean value of ten times repeated trials of random under-sampling with different sampling rate.

In addition, although under-sampling could effectively reduce the number of majority class samples, there may still exist some impurity data in majority class. In data science, we have to respect the original dataset with modifying the original dataset structure as little as possible. Hence, integrity of original majority class samples is the second objective of our proposed method. Integrity is calculated using the amount of selected majority class samples divided by the number of original majority class samples. Therefore, our proposed method is designed to solve dual objectives problems. The final dataset could obtain the highest possible Sensitivity*Specificity with the best integrity of original majority class samples.

The product of Sensitivity multiply by Specificity and the integrity of majority class are inversely in proportion. The red line and blue line in Figure 4 illustrate this situation. Therefore, there is non-unique global best solution and the suitable solutions are all recorded in a solution set, it is called non-inferior set or Pareto optimal set [36] [26]. This set contains the solutions meeting one of these conditions: 1. the solution has better performances for achieving dual objectives; 2. the solution could improve one objective and the other one does not degrade too much (1.0e-4). The decision marking of our experiment is to select a solution which produces the best Kappa statistics and Accuracy as the final results from the non-inferior set [26].

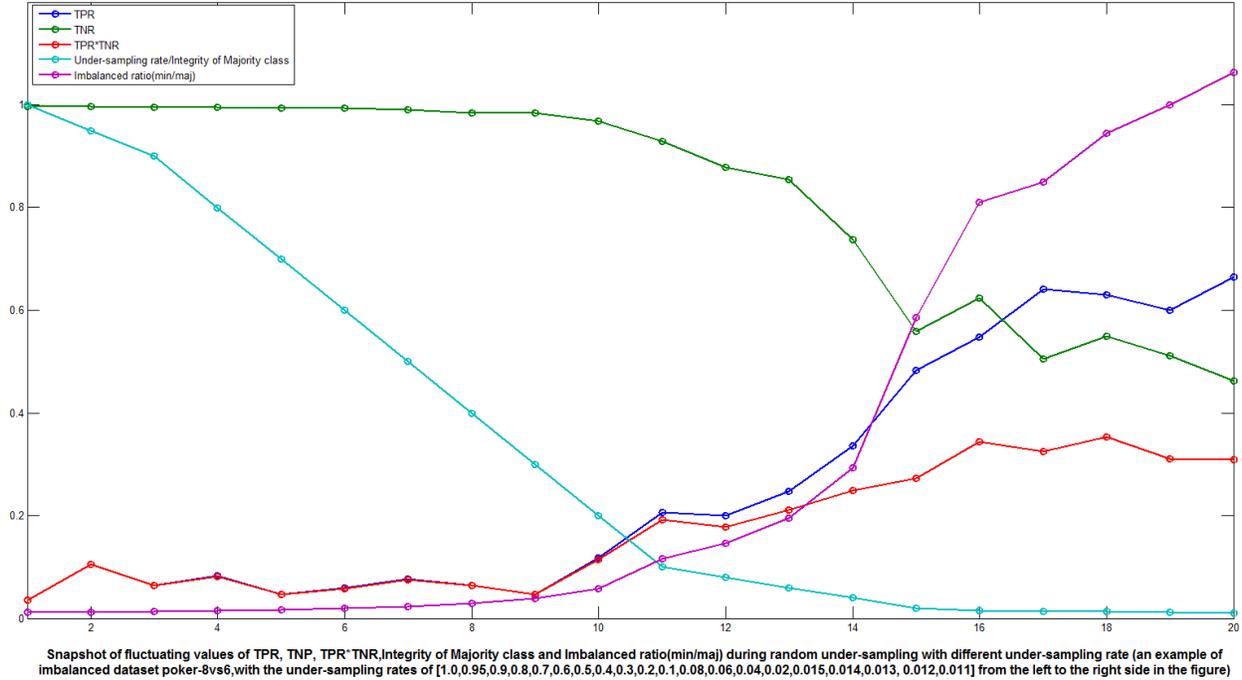

Figure 4. Snapshot of fluctuating values of TPR, TNP, TPR*TNR, Integrity of Majority class and Imbalanced ratio(min/maj) during random under-sampling with different under-sampling rate (an example of imbalanced dataset poker-8vs6,with the under-sampling rates of [1.0,0.95,0.9,0.8,0.7,0.6,0.5,0.4,0.3,0.2,0.1,0.08,0.06,0.04,0.02,0.015,0.014,0.013, 0.012,0.011] from the left to the right side in the figure)

### 3.2.4. Why choose ensemble classifier?

Ensemble techniques were mentioned in section 2, it is another main approach for solving imbalanced classification. Ensemble techniques change the tendency of classifiers to minority class through voting or assigning weights or iteration. Therefore, ensemble techniques don't need to modify the structure of the original dataset, but they indeed improve the performance of imbalanced classification. Essentially, classifying the same imbalanced dataset, ensemble classifiers are able to get better results than conventional classifiers. Furthermore, since integrity of majority class samples is one of our objectives, ensemble classifiers could maintain a higher integrity than conventional classifier even when performance of imbalanced classification for both types of classifiers are similar. That's why we adopt ensemble classifiers in our wrapper structure with BPSO. The aim of our proposed method is to improve the imbalanced classification to best performance possible while maintaining minimum damage for the original data structure through the improvement of the two main elements of the wrapper approach: searching algorithm and machine learning algorithm.

In our experiment, decision tree is selected as base learner (classifier), because it has good performance in imbalanced classification based on the experience from many other fellow researchers, for example in the special workshop in ICML-KDD 2003 [44] [10]. We respectively combined four commonly used ensemble methods of decision tree with BPSO: Bagging, AdaBoosting, Cost-sensitive and AdaCost.

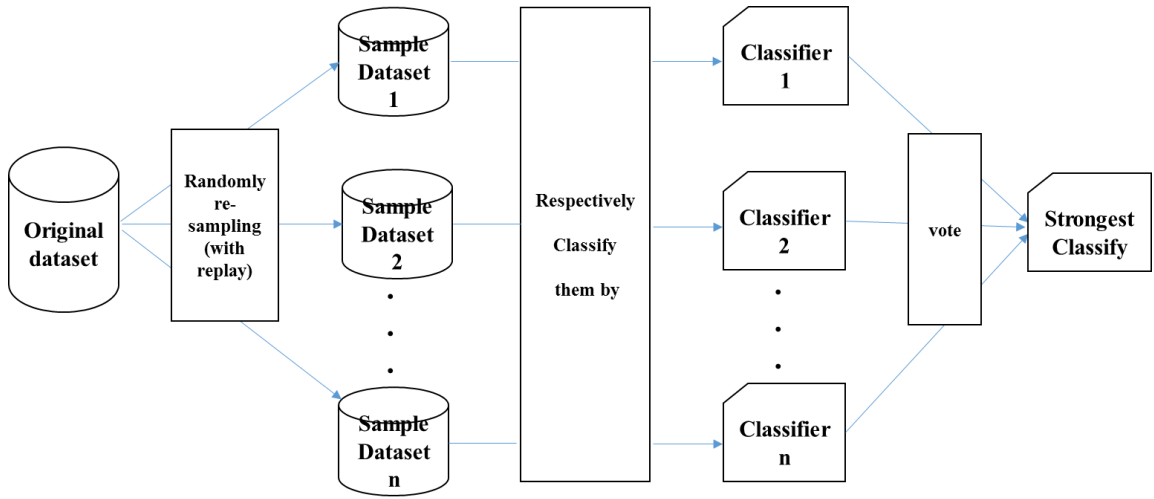

Figure 5. Principle of Bagging

The full name of Bagging is bootstrap aggregating, which is a basic ensemble learning method [39]. Its aim is to improve the performance of classification through the vote of several single classifiers, which classify the re-sampled (with replay) datasets from original dataset.

Figure 5 presents the process of bagging. The strong classifier is nominated by voting among the candidate classifies. Testing samples will be assigned to the winning classifier which gets the most votes. Bagging reduces the generalization error rate through reducing the variance of base classifier.

AdaBoosting [41] is an iteration algorithm. There will generate a new classifier on the training set in each iteration, and then the new classifier will classify all the samples to estimate the importance of each sample. It will adaptively change the distribution of samples, in order to make the base classifiers focus on those indistinguishable samples.

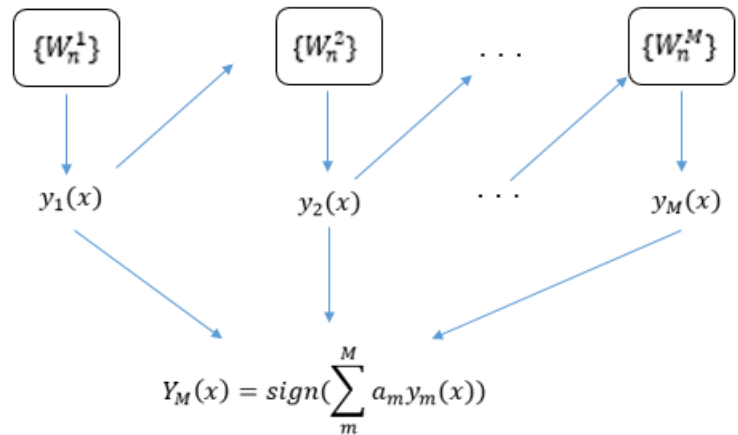

$$Y_M(x) = sign(\sum_{m}^{M} a_m y_m(x))$$

Figure 6. Flow of AdaBoosting

As shown in Figure 6, each sample in the $n$ samples of training set is assigned a weight $W_i$ ($0 \ll i < n$), which indicates the importance of each sample in classification. The weights are the same at the initial phase. By training the model in turn, the weights are constantly corrected, if the sample is classified correctly, its weight will reduce, and vice versa. The top-right corner of the oblique line in Figure 6 shows the models which are trained in turn. Therefore, the model will be more concerned with the misclassified (high weight) samples at the end of the program.

When all the programs are executed, *M* models are obtained, and they are combined into a final model $Y_M(x)$ by weighting corresponding to $y_1(x) ... y_M(x)$.

Cost-sensitive learning [11] proposes to assign different cost values to each element in the confusion matrix (Figure 3) to change the bias of classifier. Cost-sensitive technique applies a cost matrix to switch the classification problem into an optimization problem. Hence, the aim of cost-sensitive classifier is to get a classification model within minimum cost in total. Commonly, researchers set high cost for penalizing false positive class (misclassified minority class samples) and low false negative class (misclassified majority class samples) in the confusion matrix to lure the classifier concerns more about minority class samples. AdaCost is a combination method of AdaBoosting and Cost-sensitive learning [15]. The researcher of literature [15] added the value of the cost into the evaluation indexes of AdaBoosting to reduce the cost of classifiers and improve the imbalanced classification performance.

4. **Experiments and Results discussion**

   **4.1 Benchmark datasets**

In our experiment we aim to evaluate the performance of our proposed method. The whole experiment implements stratified tenflod cross-validation as the research methodology to achieve the testing part. In our experiment, our tagert is binary class imbalanced dataset. There are ten available imbalanced datasets randomly selected from KEEL tool on the website [2] and used in the experiment to examine the performance of proposed method along with other comparison algorithms. The information of these datasets are exhibited in Table 1, where #sample stands for the total amount of instances, Maj and Min respectively denotes the number of majority class sample and minority class sample, Imb.r is the short name of imbalanced ratio, which is the ratio of majority class samples and minority class samples. The imbalanced ratios of majority class and minority class are from 2.99 to 85.88.

The methods used in our experiment are coded by Matlab 2014b. The computing platform for the entire experiment is CPU: E5-1650 V2 @ 3.50 GHz, RAM: 32 GB.

Table 1. Information of used datasets in the experiment

| Dataset | #Samples | Maj | Min | Imbalance ratio |
|---|---|---|---|---|
| abalone9-18 | 731 | 689 | 42 | 16.4 |
| cleveland-0_vs_4 | 177 | 164 | 13 | 12.62 |
| glass-0-1-4-6_vs_2 | 205 | 188 | 17 | 11.06 |
| haberman | 306 | 225 | 81 | 2.78 |
| pima | 768 | 500 | 268 | 1.87 |
| poker-8_vs_6 | 1477 | 1460 | 17 | 85.88 |
| poker-9_vs_7 | 244 | 236 | 8 | 29.5 |
| vehicle3 | 846 | 634 | 212 | 2.99 |
| winequality-red-8_vs_x6-7 | 855 | 837 | 18 | 46.5 |
| yeast-0-5-6-7-9_vs_4 | 528 | 477 | 51 | 9.35 |

   **4.2 Experiment of BPSO-Instance selection**

In the first experiment we wrapped BPSO and conventional decision tree (DT) together for selecting majority instance. The quality of BPSO Instance selection-Decision tree (BPSOIS_DT) was compared with some other traditional and commonly used basic ensemble methods:

- DT: Decision Tree directly classifier imbalanced datasets.
- Resample: it is a classical sampling method, it reduces the samples with low weight and increase the number of the samples with high weight, the total quantity of samples is unchanged.
- Bagging: In our experiment, the bagging technique bagged 50 decision trees for solving imbalanced datasets.
- Cost-sensitive: In our experiment, the costs of cost matrix are: false positive class (misclassified minority class samples) is 50, low false negative class (misclassified majority class samples) is 5, as well as true positive and true negative is 0 respectively.
- AdaBoosting: 50 decision trees with 100 iterations are used in our experiment.
- AdaCost: AdaCost implements the same parameters with Cost-sensitive and AdaBoosting.
- BPSOIS_DT: BPSO instance selection with decision tree. The population of PSO is 50, the maximum iteration is 100 and inertia weight is 0.8.

In the first experiment, we recorded Kappa statistics, Accuracy, BER, MCC, G-mean, Precision, Recall, F-measure (1), TPR×TNR, Integrity and computational time for each datasets with different method. These results are all presented from Table 6 to 15 in Appendix. Table 2 and Table 3 respectively store the average value and standard deviation of different method and different metric. Furthermore, Kappa statistics [28][30] and Accuracy are selected with the two targets from these results, in order to compare the performance of methods through the visualized radar charts.

Table 2 Average value of each evaluation index for the ten dataset with different method of experiment 1(best results highlighted in bold).

| Average | Kappa | Accuracy | BER | MCC | G-mean | Precision | Recall | F1 | TPR×TNR | Integrity | Time |
|---|---|---|---|---|---|---|---|---|---|---|---|
| DT | 0.13 | 0.85 | 0.43 | 0.14 | 0.36 | 0.66 | 0.64 | 0.65 | 0.19 | 1.00 | 0.29 |
| Bagging | 0.24 | **0.89** | 0.41 | 0.30 | 0.42 | 0.91 | **0.96** | **0.93** | 0.21 | 1.00 | 2.49 |
| Resample | 0.25 | 0.86 | 0.38 | 0.25 | 0.53 | 0.91 | 0.91 | 0.91 | 0.30 | 1.00 | 0.31 |
| Cost-sensitive | 0.10 | 0.66 | 0.39 | 0.14 | 0.43 | 0.85 | 0.59 | 0.65 | 0.26 | 1.00 | 0.32 |
| AdaBoosting | 0.26 | **0.89** | 0.39 | 0.31 | 0.44 | 0.91 | **0.96** | 0.93 | 0.23 | 1.00 | 2.44 |
| AdaCost | 0.24 | 0.78 | 0.33 | 0.28 | 0.56 | **0.96** | 0.73 | 0.79 | 0.36 | 1.00 | 2.60 |
| BPSOIS_DT | **0.51** | **0.89** | **0.25** | **0.51** | **0.68** | 0.70 | 0.69 | 0.69 | **0.53** | 0.77 | 1592.60 |

Table 3. Standard deviation of each evaluation index for the ten dataset with different method of experiment 1 (best results highlighted in bold).

| Standard deviation | Kappa | Accuracy | BER | MCC | G-mean | Precision | Recall | F1 | TPR×TNR | Integrity | Time |
|---|---|---|---|---|---|---|---|---|---|---|---|
| DT | 0.12 | 0.12 | 0.06 | 0.12 | 0.25 | 0.35 | 0.37 | 0.36 | **0.15** | 0.00 | 0.05 |
| Bagging | 0.14 | 0.10 | 0.06 | 0.14 | 0.20 | **0.08** | 0.06 | 0.07 | **0.15** | 0.00 | 0.27 |
| Resample | **0.11** | 0.11 | **0.05** | **0.11** | **0.12** | **0.08** | 0.08 | 0.08 | 0.12 | 0.00 | 0.02 |
| Cost-sensitive | **0.11** | 0.27 | 0.10 | 0.13 | 0.27 | 0.29 | 0.38 | 0.37 | 0.21 | 0.00 | 0.02 |
| AdaBoosting | 0.15 | 0.09 | 0.07 | 0.14 | 0.20 | 0.08 | **0.05** | **0.06** | 0.16 | 0.00 | 0.71 |
| AdaCost | 0.16 | 0.21 | 0.10 | 0.16 | 0.23 | 0.04 | 0.31 | 0.26 | 0.20 | 0.00 | 0.09 |
| BPSOIS_DT | 0.23 | **0.09** | 0.11 | 0.23 | 0.24 | 0.19 | 0.17 | 0.17 | 0.22 | 0.21 | 894.17 |

It is easy to find that the accuracy of original imbalance classification model is high while the products of TPR and TNR are very low. Especially when the imbalanced ratio of original dataset is high, the value of accuracy gains on 1.00 but the value of TPR×TNR are zero, some metrics even reflect a negative value, like Kappa statistic. The classifications of these imbalanced datasets were improved by different methods. As a whole, the performances and improvements of conventional methods are different, under the premise of not changing the original dataset structure, these methods could improve the effect of different methods, besides Cost-sensitive learning. Since the imbalanced ratio of imbalanced datasets are not the same, the values of the elements in cost matrix are hard to be specifically assigned. Moreover, the results are not able to express which ensemble method is the best by the improvements of each metrics in Table 3. And Table 2 shows the standard deviation of these methods in experiment 1. However, it can be easily observed that the proposed BPSOIS_DT is able to enhance most of the indicators with the minimal decreasing number of majority class samples through the wrapper approach, which needs more computational time. That is due to the searching strategy and algorithm structure of wrapper approach that searching algorithms needs to constantly call the base learner to calculate the fitness function and compare each particle's results.

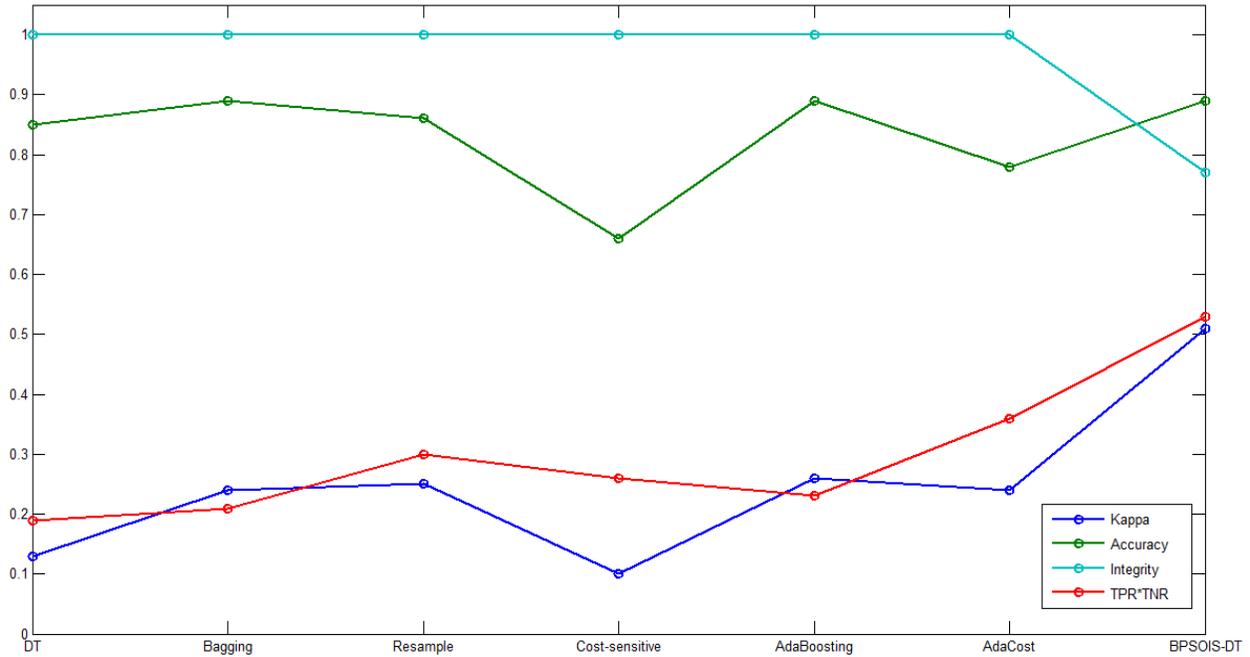

Figure 7. Average value of Kappa, Accuracy, TRP×TNR and Integrity of different methods in experiment 1

Although accuracy loses its function in imbalanced classification model, it is still an important index to appraise the classification model while the imbalanced problem is being fixed. Moreover, Kappa statistic is able to indirectly and objectively response to the robustness and credibility of classification model through the degrees of interpretation of a Kappa outcome between -1 and 1 [28]. Therefore, the average results of these two metrics with the two objectives of the proposed methods are visualized in Figure 7. It is noticeable that the cycles of Kappa, Accuracy and TPR*TNR all point to the proposed method. Combined with Table 2 for comparison that Bagging, AdaBoosting and BPSO_DT could obtain a higher accuracy, the improvement of Kappa from the two former are little and the proposed method enhance a great deal of Kappa statistics. Then the average values of resample show that it is worse than the previous three but better than AdaCost, it's worth noting that in Table 11 resample achieved the best results for the imbalanced dataset with highest imbalanced ratio while the other methods are useless.

### 4.3 Experiment of wrapping BPSO-Instance selection and ensemble methods.

The purpose of second experiment is to explore the quality of proposed BPSO Instance selection with ensemble classifiers. In addition, we simultaneously used the combination of traditional PSO with these ensemble classifiers to compare the performance of BPSO for this discrete binary problem. Since using cost-sensitive alone in experiment 1 performed unsatisfactory results in the second experiment we didn't use cost-classifier. In this round of experiment, PSO and BPSO were respectively combined with AdaBoosting, AdaCost and Bagging to implement the wrapper approach. Furthermore, three state-of-the-art under-sampling based ensemble algorithms: Easy Ensemble [29], Balance Cascade [29] and RUSboost [43], are selected for comparison. The three methods adopted 100 iterations, and the under sampling rate of RUSboos is 75%, which is a commonly used value. The same metrics in terms of

Kappa statistics, Accuracy, BER, MCC, G-mean, Precision, Recall, F-measure (1), TPR×TNR, Integrity and computational time for each datasets with different method are still recorded in the Appendix from Table 16 to Table 25, respectively.

Table 4. Average value of each evaluation index for the ten dataset with different method of experiment 2 (best results highlighted in bold).

| Average | Kappa | Accuracy | BER | MCC | G-mean | Precision | Recall | F1 | TPR×TNR | Integrity | Time |
|---|---|---|---|---|---|---|---|---|---|---|---|
| DT | 0.13 | 0.85 | 0.43 | 0.14 | 0.36 | 0.66 | 0.64 | 0.65 | 0.19 | 1.00 | 0.29 |
| EasyEnsemble | 0.23 | 0.72 | 0.27 | 0.29 | 0.73 | 0.25 | 0.75 | 0.34 | 0.54 | 1.00 | 28.64 |
| Balance Cascade | 0.25 | 0.70 | 0.26 | 0.30 | 0.73 | 0.28 | 0.77 | 0.36 | 0.54 | 1.00 | 31.28 |
| RUSboost | 0.23 | 0.86 | 0.39 | 0.23 | 0.50 | 0.35 | 0.30 | 0.30 | 0.27 | 0.75 | 12.92 |
| PSOIS_DT | 0.32 | 0.74 | 0.34 | 0.33 | 0.64 | 0.79 | 0.79 | 0.79 | 0.41 | 0.25 | 1467.82 |
| PSOIS_adab | 0.45 | 0.79 | 0.28 | 0.45 | 0.66 | 0.81 | 0.86 | 0.84 | 0.49 | 0.25 | 12869.90 |
| PSOIS_adac | 0.40 | 0.67 | 0.27 | 0.44 | 0.69 | 0.91 | 0.60 | 0.68 | 0.50 | 0.38 | 12928.91 |
| PSOIS_bagging | 0.49 | 0.78 | 0.26 | 0.50 | 0.72 | 0.79 | 0.84 | 0.81 | 0.53 | 0.21 | 11668.56 |
| BPSOIS_DT | 0.51 | **0.89** | 0.25 | 0.51 | 0.68 | 0.70 | 0.69 | 0.69 | 0.53 | 0.78 | 1592.60 |
| BPSOIS_adab | 0.57 | 0.88 | 0.24 | 0.58 | 0.73 | 0.89 | 0.86 | 0.86 | 0.54 | 0.76 | 16399.64 |
| BPSOIS_adac | 0.40 | 0.81 | 0.27 | 0.45 | 0.64 | **0.97** | 0.77 | 0.82 | 0.48 | **0.88** | 15347.72 |
| BPSOIS_bagging | **0.61** | 0.88 | **0.23** | **0.63** | **0.75** | 0.89 | **0.94** | **0.91** | **0.57** | 0.69 | 11192.47 |

Table 5. Average Standard deviation of each evaluation index for the ten dataset with different method of experiment 2 (best results highlighted in bold).

| Standard deviation | Kappa | Accuracy | BER | MCC | G-mean | Precision | Recall | F1 | TPR×TNR | Integrity | Time |
|---|---|---|---|---|---|---|---|---|---|---|---|
| DT | 0.13 | 0.13 | 0.07 | 0.13 | 0.26 | 0.37 | 0.39 | 0.38 | 0.16 | **0.00** | **0.05** |
| EasyEnsemble | 0.16 | 0.08 | 0.08 | 0.16 | 0.08 | 0.19 | 0.13 | 0.21 | **0.11** | **0.00** | 24.46 |
| Balance Cascade | 0.18 | 0.15 | 0.08 | 0.16 | 0.09 | 0.20 | 0.11 | 0.21 | 0.12 | **0.00** | 23.76 |
| RUSboost | 0.15 | 0.09 | 0.06 | 0.15 | 0.13 | 0.22 | 0.15 | 0.18 | 0.13 | **0.00** | 8.03 |
| PSOIS_DT | **0.09** | 0.08 | **0.04** | **0.09** | **0.07** | 0.10 | 0.10 | 0.10 | 0.08 | 0.15 | 60.68 |
| PSOIS_adab | 0.20 | 0.06 | 0.10 | 0.21 | 0.23 | 0.07 | 0.07 | **0.06** | 0.20 | 0.19 | 952.69 |
| PSOIS_adac | 0.26 | 0.19 | 0.13 | 0.25 | 0.18 | 0.07 | 0.27 | 0.23 | 0.22 | 0.39 | 399.16 |
| PSOIS_bagging | 0.15 | **0.06** | 0.08 | 0.14 | 0.10 | 0.06 | 0.08 | **0.06** | 0.14 | 0.18 | 1283.18 |
| BPSOIS_DT | 0.23 | 0.09 | 0.11 | 0.23 | 0.24 | 0.19 | 0.17 | 0.17 | 0.22 | 0.21 | 894.17 |
| BPSOIS_adab | 0.12 | 0.09 | 0.07 | 0.12 | 0.10 | 0.07 | 0.20 | 0.16 | 0.14 | 0.19 | 5470.77 |
| BPSOIS_adac | 0.24 | 0.20 | 0.13 | 0.23 | 0.26 | **0.03** | 0.29 | 0.24 | 0.25 | 0.15 | 5483.35 |
| BPSOIS_bagging | 0.13 | 0.09 | 0.06 | 0.13 | **0.07** | 0.08 | **0.07** | 0.07 | 0.12 | 0.23 | 1092.49 |

Performance evaluation index of these methods from the overall results of experiment 2 are displayed in the last 10 Tables in Appendix (from table 16 to 25). Wrapping BPSO Instance selection and decision tree are able to get the best results of imbalanced classification from experiment 1. Ensemble classifiers replaced signal classifier to combine with swarm intelligence algorithms in the second experiment, moreover, traditional PSO was added as comparison and it used TPR×TNR as its signal objective function since signal objective is commonly used in swarm intelligence algorithm to implement optimization. Essentially, swarms are able to find better solutions while the instance selection under-sampling technique ignores the integrity of majority class samples and only pay attention to the classification performance. However, BPSO_DT can overcome the methods, which are combined by PSO Instance selection and ensemble classifiers, not to mention the single classifier of decision tree. In wrapper approach, when BPSO Instance selection uses ensemble classifiers to replace single classifier, the power of imbalanced classification is significantly enhanced. The average results and standard deviation are demonstrated in Table 4 and Table 5 to show the effectiveness of this methods. It's noted in Table 7, Table 19 and Table 20 that when the combination of BPSO Instance selection and decision tree obtained the same performances of classification model with the unity of BPSO Instance selection and ensemble classifier, the latter maintains better integrity of original dataset. Moreover, tackling the highest imbalanced ratio dataset, poker-8_vs_6, when BPSOIS_DT and BPSOIS_adab selected almost the same amount of majority class samples, ensemble classifier achieved a remarkable performance. The statement in Section 3.2.3 is validated by these results.

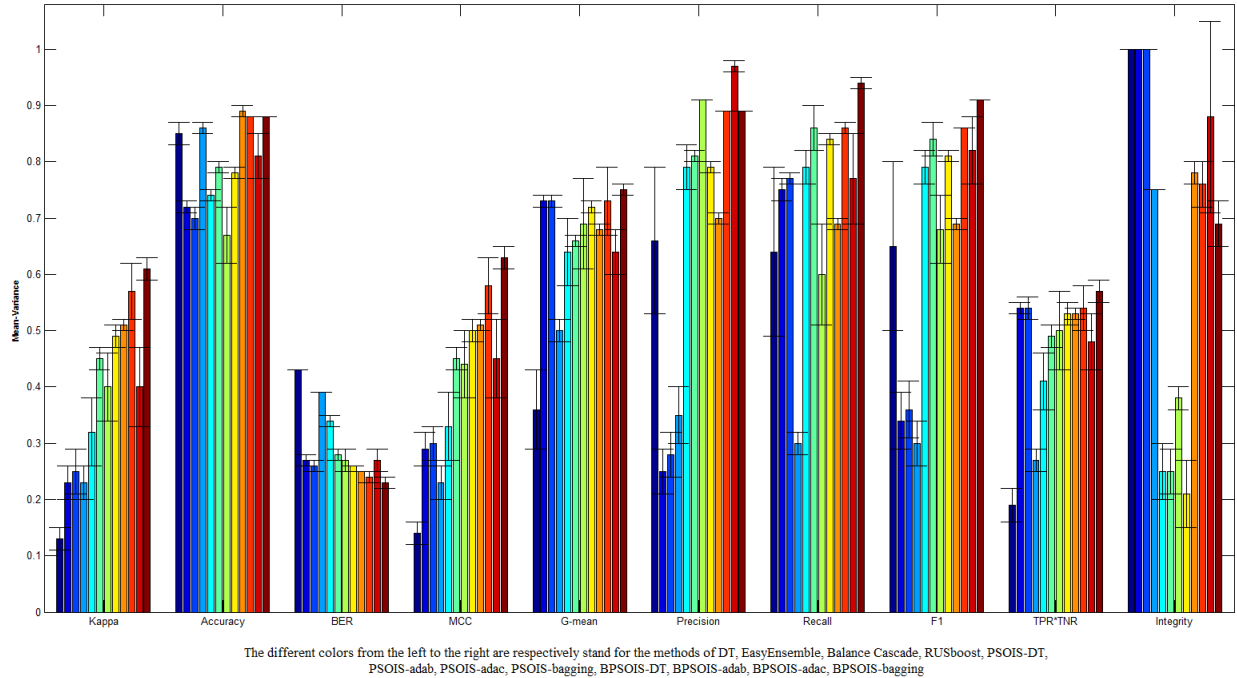

Figure 8. Mean-Variance diagram of each evaluation index for the ten dataset with different method of experiment 2

Figure 8 presents the Mean-Variance value of each evaluation index for the ten datasets with different methods of experiment 2. The different colors from the left to the right respectively represent the methods of DT, EasyEnsemble, Balance Cascade, RUSboost, PSOIS_DT, PSOIS_adab, PSOIS_adac, PSOIS_bagging, BPSOIS_DT, BPSOIS_adab, BPSOIS_adac, BPSOIS_bagging. Variance values reflect the fluctuations of different performance of different methods. The three state-of-the-art algorithms obtained similar effects, which are better than the basic methods in experiment 1. However, although they could get relatively higher value of the product of TPR and TNR, their general performances are still not as good as our proposed methods. Besides Easy Ensemble and Balance Cascade, BPSOIS with AdaCost maintained the most number of original samples, but its classification performance is not the best. The other three cycles are obviously biased towards BPSOIS_bagging which obtained the best average performances with lower time cost shown in Table 4. BPSOIS_adab is slightly better than BPSOIS_DT in general. Since the elements in the cost matrix are difficult to be specifically assigned, cost-sensitive learning is not able to play its function, it might even have bad influence on AdaCost algorithm. Therefore, BPSOIS_bagging is able to attain the best results in all of these methods. Figure 9 to Figure 18 respectively illustrate the final non-inferior of BPSOIS_bagging for processing each imbalanced dataset in the experiment. We can find that in these figures, they contain more than one solutions, besides poker-9_vs_7. In the searching process, these solutions attempted to get closer to the optimal point at the upper right corner in the two dimensional coordinates. These solutions making up the surface is called Pareto surface [36], they are all the possible best solutions. This experiment adopted the decision making that whichever solution can produce the best product of kappa, and accuracy is the ultimate result.

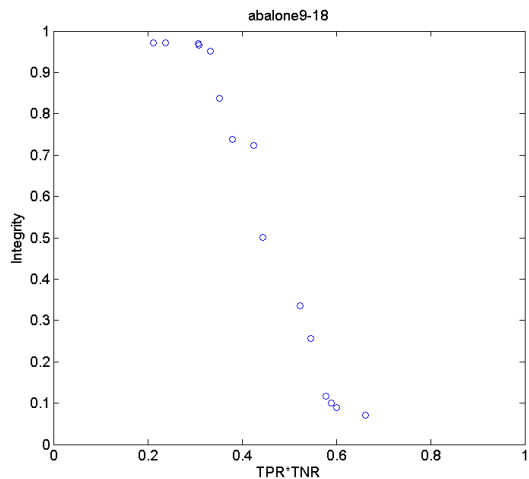
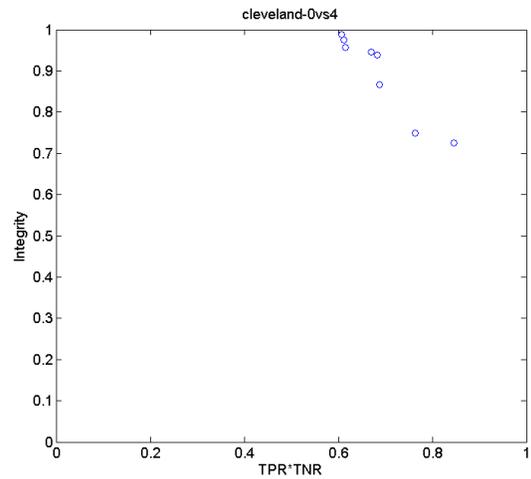

Figure 9. Non-inferior set of BPSOIS_bagging for abalone 9-18      Figure 10. Non-inferior set of BPSOIS_bagging for cleveland-0_vs_4

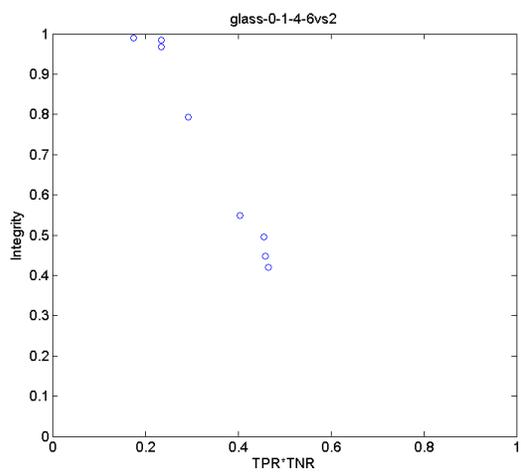
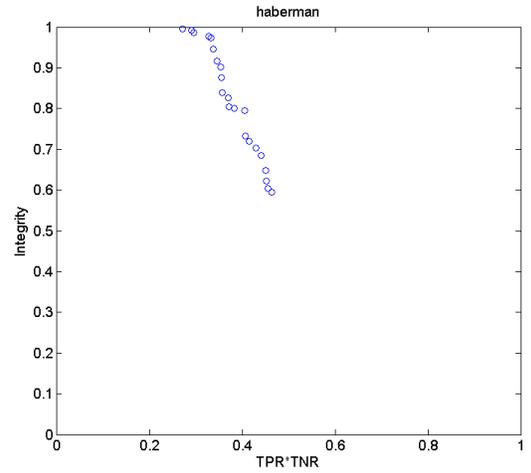

Figure 11. Non-inferior set of BPSOIS_bagging for glass-0-1-4-6_vs_2      Figure 12. Non-inferior set of BPSOIS_bagging for haberman

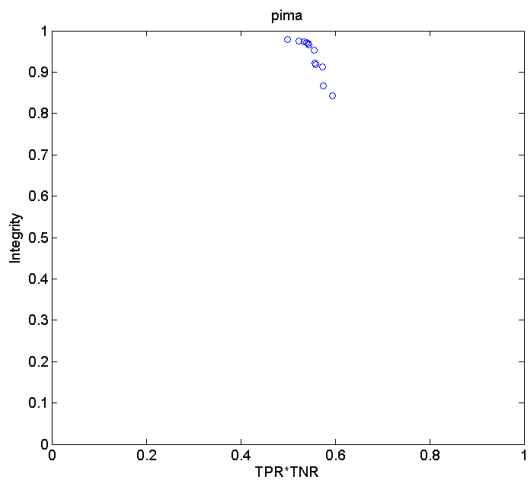
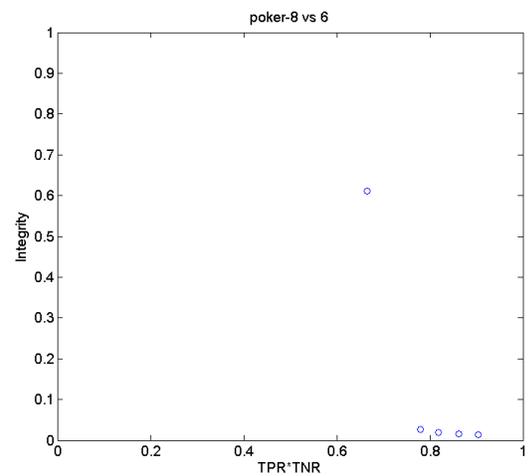

Figure 13. Non-inferior set of BPSOIS_bagging for pima      Figure 14. Non-inferior set of BPSOIS_bagging for poker-8_vs_6

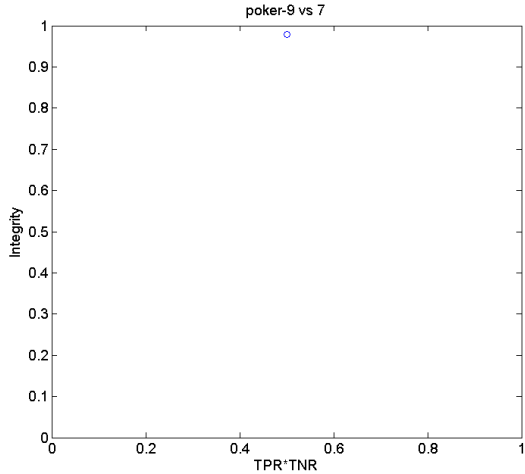
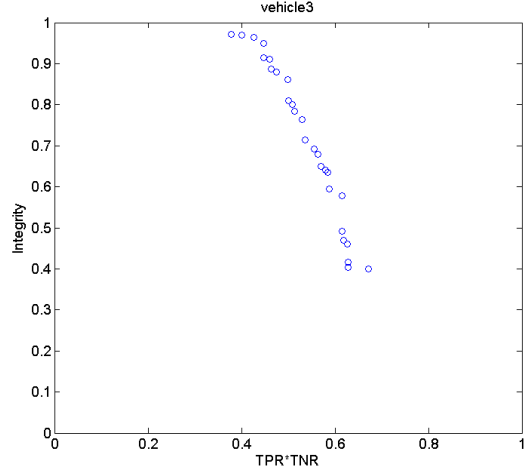

Figure 105. Non-inferior set of BPSOIS_bagging for poker-9_vs_7     Figure 16. Non-inferior set of BPSOIS_bagging for vehicle 3

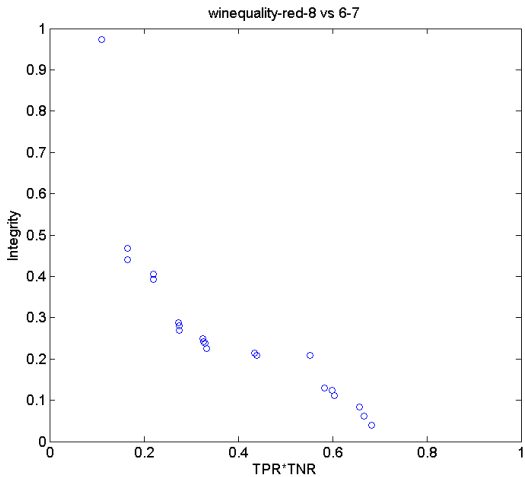
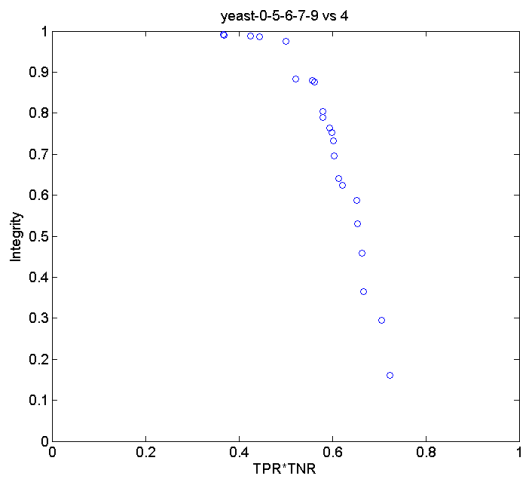

Figure 17. Non-inferior set of BPSOIS_bagging for winequality-red -8_vs_6-7 9-18     Figure 18. Non-inferior set of BPSOIS_bagging for yeast_0-5-6-7 -9_vs_4

## 5. Conclusion

This paper proposed a new ensemble method to solve imbalanced classification problems. The proposed method absorbs the advantages of under-sample and ensemble learning through wrapping a new BPSO instance selection with ensemble classifiers. BPSOIS is a multi-objective algorithm, it maximally improves the classification performance while minimizes the damage to the integrity of original samples in dataset through constantly and trajectory searching for particles. Furthermore, BPSO transforms the NP hard problem of instance selection into a binary function optimization problem to improve the efficiency and correctness. The first experimental results obviously show that, BPSOIS has significantly better performance comparing to the other conventional basic ensemble methods experimented, while it adopts single classifier as base learner. In addition, the proposed method outperformed three state-of-the-art under-sampling methods, validated by experimental results, and the performances of BPSO are also comprehensively better than PSO in whole. The results of second experiment illustrate the power and effectiveness of the combination of BPSOIS and ensemble classifiers and the statement in Section 3.2.3 is validated. Experiments show that, the ensemble of BPSOIS and Bagging could obtain the best results of all.

# Acknowledgement

The authors are grateful for the financial support from the Research Grants, (1) Nature-Inspired Computing and Meta-heuristics Algorithms for Optimizing Data Mining Performance, Grant no. MYRG2016-00069-FST, offered by the University of Macau, FST, and RDAO; and (2) A Scalable Data Stream Mining Methodology: Stream-based Holistic Analytics and Reasoning in Parallel, Grant no. FDCT/126/2014/A3, offered by FDCT Macau.

# Appendix

## Results of Experiment 1.

Table 6. Results of abalone9-18 with different method in experiment 1

| abalone9-18 | Kappa | Accuracy | BER | MCC | G-mean | Precision | Recall | F1 | TPR*TNR | Integrity | Time |
|---|---|---|---|---|---|---|---|---|---|---|---|
| DT | 0.23 | 0.91 | 0.38 | 0.23 | 0.52 | 0.96 | 0.95 | 0.95 | 0.27 | 1.00 | 0.33 |
| Resample | 0.28 | 0.91 | 0.35 | 0.28 | 0.58 | 0.96 | 0.95 | 0.95 | 0.34 | 1.00 | 0.32 |
| Bagging | 0.27 | 0.95 | 0.41 | 0.32 | 0.43 | 0.95 | 0.99 | 0.97 | 0.19 | 1.00 | 2.49 |
| Cost-sensitive | 0.18 | 0.78 | 0.27 | 0.25 | 0.72 | 0.97 | 0.79 | 0.87 | 0.53 | 1.00 | 0.37 |
| AdaBoosting | 0.28 | 0.95 | 0.41 | 0.36 | 0.44 | 0.95 | 1.00 | 0.97 | 0.19 | 1.00 | 2.66 |
| AdaCost | 0.25 | 0.86 | 0.29 | 0.28 | 0.69 | 0.97 | 0.88 | 0.92 | 0.48 | 1.00 | 2.61 |
| BPSOIS_DT | 0.44 | 0.94 | 0.28 | 0.44 | 0.68 | 0.47 | 0.48 | 0.47 | 0.46 | 0.99 | 1583.22 |

Table 7. Results of cleveland-0_vs_4 with different method in experiment 1

| cleveland-0_vs_4 | Kappa | Accuracy | BER | MCC | G-mean | Precision | Recall | F1 | TPR*TNR | Integrity | Time |
|---|---|---|---|---|---|---|---|---|---|---|---|
| DT | 0.16 | 0.91 | 0.44 | 0.17 | 0.39 | 0.29 | 0.15 | 0.20 | 0.15 | 1.00 | 0.18 |
| Resample | 0.44 | 0.93 | 0.29 | 0.44 | 0.67 | 0.96 | 0.96 | 0.96 | 0.44 | 1.00 | 0.30 |
| Bagging | 0.42 | 0.94 | 0.35 | 0.47 | 0.55 | 0.95 | 0.99 | 0.97 | 0.31 | 1.00 | 2.23 |
| Cost-sensitive | 0.03 | 0.78 | 0.47 | 0.04 | 0.44 | 0.93 | 0.82 | 0.87 | 0.19 | 1.00 | 0.33 |
| AdaBoosting | 0.52 | 0.94 | 0.28 | 0.53 | 0.67 | 0.96 | 0.98 | 0.97 | 0.45 | 1.00 | 2.60 |
| AdaCost | 0.59 | 0.94 | 0.18 | 0.59 | 0.81 | 0.98 | 0.96 | 0.97 | 0.66 | 1.00 | 2.49 |
| BPSOIS_DT | 0.80 | 0.97 | 0.09 | 0.80 | 0.91 | 0.79 | 0.85 | 0.81 | 0.83 | 0.86 | 862.95 |

Table 8. Results of glass-0-1-4-6_vs_2 with different method in experiment 1

| glass-0-1-4-6_vs_2 | Kappa | Accuracy | BER | MCC | G-mean | Precision | Recall | F1 | TPR*TNR | Integrity | Time |
|---|---|---|---|---|---|---|---|---|---|---|---|
| DT | 0.08 | 0.88 | 0.47 | 0.09 | 0.33 | 0.18 | 0.12 | 0.14 | 0.11 | 1.00 | 0.21 |
| Resample | 0.01 | 0.87 | 0.50 | 0.01 | 0.24 | 0.92 | 0.95 | 0.93 | 0.06 | 1.00 | 0.30 |
| Bagging | 0.10 | 0.92 | 0.47 | 0.23 | 0.24 | 0.92 | 1.00 | 0.96 | 0.06 | 1.00 | 2.23 |
| Cost-sensitive | 0.08 | 0.42 | 0.34 | 0.18 | 0.60 | 0.99 | 0.38 | 0.55 | 0.36 | 1.00 | 0.30 |
| AdaBoosting | 0.20 | 0.93 | 0.44 | 0.33 | 0.34 | 0.93 | 1.00 | 0.96 | 0.12 | 1.00 | 2.61 |
| AdaCost | 0.24 | 0.80 | 0.29 | 0.28 | 0.70 | 0.96 | 0.82 | 0.89 | 0.48 | 1.00 | 2.59 |
| BPSOIS_DT | 0.37 | 0.89 | 0.30 | 0.37 | 0.66 | 0.40 | 0.47 | 0.43 | 0.44 | 0.90 | 879.26 |

Table 9. Results of haberman with different method in experiment 1

| haberman | Kappa | Accuracy | BER | MCC | G-mean | Precision | Recall | F1 | TPR*TNR | Integrity | Time |
|---|---|---|---|---|---|---|---|---|---|---|---|
| DT | 0.06 | 0.62 | 0.47 | 0.06 | 0.50 | 0.75 | 0.72 | 0.73 | 0.25 | 1.00 | 0.31 |
| Resample | 0.21 | 0.69 | 0.39 | 0.21 | 0.59 | 0.79 | 0.77 | 0.78 | 0.34 | 1.00 | 0.28 |
| Bagging | 0.16 | 0.69 | 0.43 | 0.16 | 0.52 | 0.77 | 0.83 | 0.80 | 0.27 | 1.00 | 2.37 |
| Cost-sensitive | 0.00 | 0.26 | 0.50 | 0.00 | 0.00 | 0.00 | 0.00 | 0.00 | 0.00 | 1.00 | 0.31 |
| AdaBoosting | 0.27 | 0.75 | 0.38 | 0.28 | 0.56 | 0.79 | 0.89 | 0.84 | 0.31 | 1.00 | 2.58 |
| AdaCost | 0.03 | 0.32 | 0.48 | 0.08 | 0.29 | 0.86 | 0.08 | 0.15 | 0.08 | 1.00 | 2.47 |
| BPSOIS_DT | 0.41 | 0.75 | 0.30 | 0.41 | 0.68 | 0.63 | 0.56 | 0.59 | 0.47 | 0.73 | 1226.42 |

Table 10. Results of pima with different method in experiment 1

| pima | Kappa | Accuracy | BER | MCC | G-mean | Precision | Recall | F1 | TPR*TNR | Integrity | Time |
|---|---|---|---|---|---|---|---|---|---|---|---|
| DT | 0.27 | 0.67 | 0.36 | 0.27 | 0.63 | 0.75 | 0.73 | 0.74 | 0.40 | 1.00 | 0.33 |
| Resample | 0.31 | 0.69 | 0.35 | 0.31 | 0.64 | 0.76 | 0.77 | 0.76 | 0.42 | 1.00 | 0.32 |
| Bagging | 0.44 | 0.76 | 0.29 | 0.45 | 0.70 | 0.79 | 0.86 | 0.82 | 0.49 | 1.00 | 2.81 |
| Cost-sensitive | 0.10 | 0.45 | 0.43 | 0.18 | 0.42 | 0.86 | 0.19 | 0.31 | 0.18 | 1.00 | 0.32 |
| AdaBoosting | 0.43 | 0.75 | 0.29 | 0.43 | 0.69 | 0.78 | 0.85 | 0.81 | 0.48 | 1.00 | 2.72 |
| AdaCost | 0.29 | 0.60 | 0.32 | 0.37 | 0.63 | 0.93 | 0.42 | 0.57 | 0.39 | 1.00 | 2.61 |
| BPSOIS_DT | 0.54 | 0.78 | 0.23 | 0.54 | 0.76 | 0.73 | 0.70 | 0.72 | 0.58 | 0.80 | 2614.66 |

| poker-8_vs_6 | Kappa | Accuracy | BER | MCC | G-mean | Precision | Recall | F1 | TPR*TNR | Integrity | Time |
|---|---|---|---|---|---|---|---|---|---|---|---|
| DT | 0.00 | 0.99 | 0.50 | 0.00 | 0.00 | 0.99 | 1.00 | 0.99 | 0.00 | 1.00 | 0.34 |
| Resample | 0.15 | 0.98 | 0.42 | 0.15 | 0.42 | 0.99 | 0.99 | 0.99 | 0.17 | 1.00 | 0.31 |
| Bagging | 0.00 | 0.99 | 0.50 | 0.00 | 0.00 | 0.99 | 1.00 | 0.99 | 0.00 | 1.00 | 2.61 |
| Cost-sensitive | 0.00 | 0.99 | 0.50 | 0.00 | 0.00 | 0.99 | 1.00 | 0.99 | 0.00 | 1.00 | 0.32 |
| AdaBoosting | 0.00 | 0.99 | 0.50 | 0.00 | 0.00 | 0.99 | 1.00 | 0.99 | 0.00 | 1.00 | 0.32 |
| AdaCost | 0.00 | 0.99 | 0.50 | 0.00 | 0.00 | 0.99 | 1.00 | 0.99 | 0.00 | 1.00 | 2.76 |
| BPSOIS_DT | 0.00 | 0.99 | 0.50 | 0.00 | 0.00 | 0.99 | 1.00 | 0.99 | 0.00 | 0.98 | 1840.77 |

| poker-9_vs_7 | Kappa | Accuracy | BER | MCC | G-mean | Precision | Recall | F1 | TPR*TNR | Integrity | |
|---|---|---|---|---|---|---|---|---|---|---|---|
| DT | -0.02 | 0.95 | 0.51 | -0.02 | 0.00 | 0.97 | 0.98 | 0.97 | 0.00 | 1.00 | 0.30 |
| Resample | 0.29 | 0.96 | 0.38 | 0.30 | 0.50 | 0.97 | 0.99 | 0.98 | 0.25 | 1.00 | 0.29 |
| Bagging | 0.22 | 0.97 | 0.44 | 0.35 | 0.35 | 0.97 | 1.00 | 0.99 | 0.13 | 1.00 | 2.19 |
| Cost-sensitive | 0.00 | 0.26 | 0.50 | 0.03 | 0.15 | 0.83 | 0.02 | 0.05 | 0.02 | 1.00 | 0.33 |
| AdaBoosting | 0.17 | 0.96 | 0.44 | 0.19 | 0.35 | 0.97 | 0.99 | 0.98 | 0.12 | 1.00 | 2.61 |
| AdaCost | 0.24 | 0.95 | 0.39 | 0.24 | 0.49 | 0.97 | 0.98 | 0.98 | 0.24 | 1.00 | 2.54 |
| BPSOIS_DT | 0.85 | 0.99 | 0.13 | 0.86 | 0.87 | 1.00 | 0.75 | 0.86 | 0.75 | 0.72 | 854.87 |

| vehicle3 | Kappa | Accuracy | BER | MCC | G-mean | Precision | Recall | F1 | TPR*TNR | Integrity | |
|---|---|---|---|---|---|---|---|---|---|---|---|
| DT | 0.27 | 0.73 | 0.37 | 0.27 | 0.60 | 0.81 | 0.83 | 0.82 | 0.36 | 1.00 | 0.34 |
| Resample | 0.28 | 0.73 | 0.36 | 0.28 | 0.61 | 0.82 | 0.83 | 0.82 | 0.37 | 1.00 | 0.33 |
| Bagging | 0.35 | 0.79 | 0.35 | 0.37 | 0.59 | 0.82 | 0.92 | 0.87 | 0.35 | 1.00 | 3.09 |
| Cost-sensitive | 0.12 | 0.91 | 0.35 | 0.16 | 0.60 | 0.99 | 0.92 | 0.95 | 0.36 | 1.00 | 0.32 |
| AdaBoosting | 0.27 | 0.78 | 0.39 | 0.30 | 0.50 | 0.79 | 0.95 | 0.86 | 0.25 | 1.00 | 2.82 |
| AdaCost | 0.21 | 0.52 | 0.33 | 0.32 | 0.60 | 0.97 | 0.37 | 0.53 | 0.36 | 1.00 | 2.72 |
| BPSOIS_DT | 0.48 | 0.77 | 0.27 | 0.49 | 0.72 | 0.79 | 0.86 | 0.82 | 0.53 | 0.56 | 3774.49 |

| wine quality-red-8_vs_6-7 | Kappa | Accuracy | BER | MCC | G-mean | Precision | Recall | F1 | TPR*TNR | Integrity | Time |
|---|---|---|---|---|---|---|---|---|---|---|---|
| DT | -0.02 | 0.96 | 0.51 | -0.02 | 0.00 | 0.00 | 0.00 | 0.00 | 0.00 | 1.00 | 0.30 |
| Resample | 0.17 | 0.96 | 0.40 | 0.17 | 0.47 | 0.98 | 0.98 | 0.98 | 0.22 | 1.00 | 0.30 |
| Bagging | 0.10 | 0.98 | 0.47 | 0.23 | 0.24 | 0.98 | 1.00 | 0.99 | 0.06 | 1.00 | 2.54 |